\documentclass[letterpaper, 10 pt, journal, twoside]{IEEEtran}

\usepackage{graphicx}
\usepackage{amsmath,amssymb}
\usepackage{bm}
\usepackage{multirow}

\makeatletter
\let\MYcaption\@makecaption
\makeatother

\usepackage{subcaption}
\makeatletter
\let\@makecaption\MYcaption
\makeatother

\usepackage[%
    colorlinks=true,
    pdfborder={0 0 0},
    linkcolor=blue
]{hyperref}
\usepackage{comment}

\newcommand{\argmin}{\mathop{\rm arg~min}\limits}

\usepackage{color}
\usepackage{siunitx}
\usepackage{cite}
\begin{document}

\title{Stereo-LiDAR Fusion by Semi-Global Matching with Discrete Disparity-Matching Cost and Semidensification}

\markboth{IEEE Robotics and Automation Letters. Preprint Version. Accepted March, 2025}
{Yao \MakeLowercase{\textit{et al.}}: Stereo-LiDAR Fusion by Semi-Global Matching with Discrete Disparity-Matching Cost and Semidensification} 

\author{Yasuhiro~Yao,
        Ryoichi~Ishikawa,~\IEEEmembership{Member, IEEE},
        and~Takeshi~Oishi,~\IEEEmembership{Member, IEEE}

\thanks{Manuscript received: October, 22, 2024; Revised February, 3, 2025; Accepted March, 3, 2025.}
\thanks{This paper was recommended for publication by Editor Cadena Lerma, Cesar upon evaluation of the Associate Editor and Reviewers' comments.
This work was partially supported by JSPS KAKENHI Grant Numbers JP24K21173 and JP24H00351.} 
\thanks{Y.\ Yao, R.\ Ishikawa, and T.\ Oishi were with the Institute of Industrial Science, University of Tokyo, Tokyo, 153-8505 Japan (e-mail: yao@cvl.iis.u-tokyo.ac.jp)}
}

\maketitle

\begin{abstract}
We present a real-time, non-learning depth estimation method that fuses Light Detection and Ranging (LiDAR) data with stereo camera input. 
Our approach comprises three key techniques: Semi-Global Matching (SGM) stereo with Discrete Disparity-matching Cost (DDC), semidensification of LiDAR disparity, and a consistency check that combines stereo images and LiDAR data. 
Each of these components is designed for parallelization on a GPU to realize real-time performance. 
When it was evaluated on the KITTI dataset, the proposed method achieved an error rate of 2.79\%, outperforming the previous state-of-the-art real-time stereo-LiDAR fusion method, which had an error rate of 3.05\%. 
Furthermore, we tested the proposed method in various scenarios, including different LiDAR point densities, varying weather conditions, and indoor environments, to demonstrate its high adaptability. 
We believe that the real-time and non-learning nature of our method makes it highly practical for applications in robotics and automation. 
\end{abstract}


\section{Introduction}
\IEEEPARstart{R}{eal}-time depth estimation is crucial for a wide range of robotics and automation applications. 
Depth data are needed not only by autonomous vehicles but also by various mobile systems and robots to understand and navigate their environment. 
Standard methods for depth measurements include triangulation and time of flight (ToF). 
The widely used devices for these methods are stereo cameras, which rely on triangulation, and Light Detection And Ranging (LiDAR), which operates using ToF principles. 

Both stereo cameras and LiDAR systems have distinct advantages and limitations. 
Stereo cameras deliver depth information with a high resolution that is equivalent to the resolution of the input images. 
However, they perform poorly on untextured surfaces, repetitive patterns, and low-light environments due to challenges in finding correspondences. 
In contrast, LiDAR provides more precise depth measurements and is robust against variations in lighting and surface texture. 
However, LiDAR data are sparse because LiDAR captures depth information only at specific points where the laser beam intersects the target scene. 

Sensor fusion overcomes these drawbacks of sensor systems. 
In particular, a stereo-LiDAR fusion system can obtain a highly accurate, high-resolution depth map without being affected by the environment or by scenes~\cite{yao2021non}. 
As with other research topics, sensor fusion systems employ a learning-based~\cite{ park2018high, wang20193d, park2019high, cheng2019noise, zhang2020listereo, meng2021gpu,choe2021volumetric, zhang2022slfnet, meng2023fastfusion} or non-learning-based~\cite{huber2011integrating, maddern2016real, yao2021non, forkel2023lidar} strategy. 
For stereo-LiDAR fusion, the performance of learning-based methods can be domain-dependent, as training is conducted on a specific dataset. 
Meanwhile, non-learning methods have the advantage of being less dependent on specific datasets and domains. 

\begin{figure}[t]
\centering
\includegraphics[width=\linewidth]{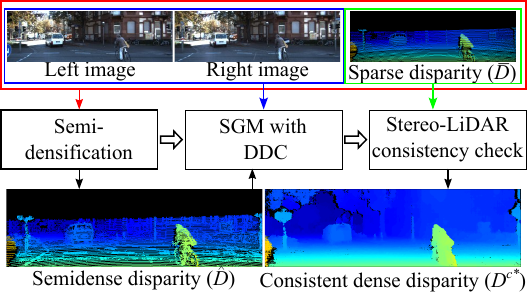}
\caption{Flow chart of the proposed method. 
The semidensification process takes stereo images and the sparse disparity map ($\bar{D}$) and outputs the semidense disparity map ($\hat{D}$). 
SGM with DDC takes stereo images with either a semidense disparity map ($\hat{D}$) or a sparse disparity map ($\bar{D}$) and outputs a dense disparity map. 
The stereo-LiDAR consistency check annotates invalid disparities based on the consistency of the three views to obtain a consistent dense disparity map ($D^{c*}$).}
\label{fig:flow}
\end{figure}

However, the previous state-of-the-art real-time non-learning methods are not robust due to outlier-sensitive costs and direct use of LiDAR disparities, including misprojections~\cite{forkel2023lidar}. 
Therefore, we propose a stereo-LiDAR fusion method, particularly a non-learning approach that operates in real time and achieves an accuracy comparable to that of learning-based methods. 
Fig.~\ref{fig:flow} is an overview of the proposed method. 
We employ Semi-Global Matching (SGM)~\cite{hirschmuller2007stereo} as the base stereo algorithm. 
The primary reason for the suboptimal accuracy of stereo-LiDAR fusion is that the integration between the stereo camera and LiDAR is insufficient. 
The proposed method addresses this issue by introducing the following three key approaches: 
\begin{itemize}
    \item Discrete Disparity-matching Cost (DDC) discretely evaluates sparse disparities in the SGM framework. 
    \item Semidensification partially densifies sparse disparities to provide prior information to SGM using DDC. 
    \item Stereo-LiDAR consistency check ensures consistency in the disparity estimation by leveraging three views from the stereo cameras and LiDAR. 
\end{itemize}
In addition, we demonstrate that the proposed method surpasses previous state-of-the-art (SOTA) real-time stereo-LiDAR fusion techniques and exhibits strong adaptability across various domains. 

Sec.~\ref{sec:related_work} reviews related works to position the proposed methodology within the existing literature. 
In Sec.~\ref{sec:SGM}, we provide a brief overview of stereo SGM to facilitate a better understanding of the proposed approach and the variable notation. 
Secs.~\ref{sec:method_cost}, \ref{sec:method_semi}, and \ref{sec:method_consistency} describe DDC, semidensification, and the stereo-LiDAR consistency check, respectively. 
We evaluate the performance of the proposed method in Sec.~\ref{sec:eval} and present the conclusions in Sec.~\ref{sec:conclusion}. 

%

\section{Related Work}
\label{sec:related_work}
We consider the stereo system to be a parallel dense stereo setup, in which disparity maps are generated from a pair of images captured by two cameras aligned parallel to their image planes. 
Although modern deep learning methods estimate relative depth from a monocular image\cite{yang2024depth}, stereo images are still required to obtain depth at the real scale, which is our subject matter. 
In the following sections, we present a brief overview of related work on parallel dense stereo and stereo-LiDAR fusion methods. 

\subsection{Parallel dense stereo}

The stereo system estimates disparities by analyzing the local similarity between two images along their epipolar lines. 
A widely used method for finding the optimal solution is the energy minimization approach~\cite{boykov2001fast, felzenszwalb2006efficient,hirschmuller2007stereo}, which includes pixel-matching cost and smoothness terms in 2D space. 
In most cases, this minimization problem is considered NP-hard~\cite{boykov2001fast}. 
Although several methods, such as graph cuts~\cite{boykov2001fast} and belief propagation~\cite{felzenszwalb2006efficient}, have been proposed to solve this problem, these methods are computationally expensive. 

In contrast, SGM~\cite{hirschmuller2007stereo} reduces computational costs by approximating the 2D smoothness constraint using a combination of multiple one-dimensional (1D) constraints. 
Currently, SGM is one of the most widely used stereo-matching methods due to its high performance. 
Several variants of SGM have also been developed~\cite{yamaguchi2014efficient, hernandez2016embedded, kallwies2020triple}. 
Learning-based dense stereo techniques have recently been introduced~\cite{wang2021pvstereo, guo2023openstereo}; however, training a model to handle all potential and unforeseen scenarios remains a challenge. 
For this reason, we consider SGM to be a suitable foundational algorithm for stereo matching. 

\subsection{Stereo-LiDAR fusion}
Non-learning stereo-LiDAR fusion has evolved and improved over the years. 
Badino et al.\ utilized LiDAR data to narrow the stereo matching search space and introduced predefined paths for dynamic programing~\cite{huber2011integrating}. 
Maddern et al.\ proposed a probabilistic model to fuse LiDAR and stereo disparities by combining the priors of individual sensors~\cite{maddern2016real}. 
Yao et al.\ proposed a method for selecting, using belief propagation, appropriate depth values from LiDAR projections in the surrounding area~\cite{yao2021non} and then smoothing using total generalized variation~\cite{yao2020discontinuous}. 
Forkel et al.\ incorporated a LiDAR-based matching cost into SGM stereo to determine whether an estimated depth was similar to the LiDAR measurement of the depth~\cite{forkel2023lidar}. 

Many recent studies have adopted learning-based approaches and have led to significant improvements in accuracy. 
Park et al.\ first developed a neural network (NN) that integrated LiDAR and stereo disparities~\cite{park2018high}, and they formulated the problem of uncalibrated sensor fusion in a unified deep learning framework~\cite{park2019high}. 
Wang et al.\ employed a stereo-matching network with enhanced techniques rather than directly fusing estimated depths across LiDAR and stereo modalities~\cite{wang20193d}. 
Cheng et al.\ proposed a self-supervised method for training an NN to remove occluded LiDAR projections, enabling the inference of dense disparity maps~\cite{cheng2019noise}. 
Choe et al.\ introduced a geometry-aware network for long-range depth estimation~\cite{choe2021volumetric}. 
Zhang et al.\ proposed a method for coupling depth cues in two modalities in a compact network architecture~\cite{zhang2022slfnet}. 
Meng et al.\ presented a real-time, NN-based approach for coarse depth prediction and subsequent depth refinement~\cite{meng2021gpu, meng2023fastfusion}. 

Among the comparison methods, only~\cite{maddern2016real,huber2011integrating} and~\cite{forkel2023lidar} meet the criteria for real-time processing and do not require learning. 
However, these approaches are less accurate than offline or learning-based methods. 
In contrast, the proposed method is a real-time, non-learning approach that achieves competitive accuracy with both offline and learning-based methods.

\section{Preliminary: Stereo SGM}
\label{sec:SGM}
In this section, we present an overview of stereo SGM~\cite{hirschmuller2007stereo}. 
SGM utilizes a strategy that minimizes the cost of pixel-wise matching while applying smoothness constraints to estimate the disparity image. 
We define the matching cost for a pixel $\mathbf{p} \in \Omega$ at a possible disparity $d_\mathbf{p}\in \mathbb{N}$ as $C(\mathbf{p}, d_\mathbf{p})$, where $\Omega \subset \mathbb{N}^2$ is the set of pixel coordinates. 
Relying solely on matching costs may result in inconsistencies across the disparity map $D = \left\{ d_\mathbf{p} \mid \mathbf{p} \in \Omega \right\}$. 
To address this problem, SGM introduces a smoothness term that penalizes significant changes in disparity between neighboring pixels as follows: 
\begin{align}    
    E(D) =&     \sum_\mathbf{p}{\bigg\lbrace C\left(\mathbf{p}, d_\mathbf{p}\right)}\ + \sum_{\mathbf{q}\in N_\mathbf{p}}{P_1T\bigl[\left|d_\mathbf{p} - d_\mathbf{q}\right| = 1\bigr]} \nonumber \\
    &+ \sum_{\mathbf{q}\in N_\mathbf{p}}{P_2T\bigl[\left|d_\mathbf{p} - d_\mathbf{q}\right| > 1\bigr]} \bigg\rbrace,
    \label{eq:energy}
\end{align}
where $T[\cdot]=1$ when $\cdot$ is true, and $T[\cdot]=0$ otherwise. 
$N_\mathbf{p}$ represents the neighboring pixels of pixel $\mathbf{p}$. 
$P_1$ is a constant penalty applied to neighboring pixel $\mathbf{q} \in N_\mathbf{p}$ when there is a small change in disparity (i.e., by one pixel). 
$P_2$ is a constant penalty for large changes in disparity. 
We obtain the optimal disparity image $D^{\ast} = \left\{d_\mathbf{p}^{\ast} \mid \mathbf{p} \in \Omega \right\}$ by minimizing $E(D)$. 
Such global minimization in 2D is NP-complete for many discontinuity-preserving energies~\cite{boykov2001fast}. 

SGM divides the problem into several 1D paths in the image. 
We used vertical, horizontal, and diagonal paths (eight paths in total) in this study. 
The cost $L_{\bf r}$ of each path ${\bf r}$ is calculated by the propagation along with the path ${\bf r}$ as: 
\begin{align}
    L_{\bf r} (\mathbf{p}, d_\mathbf{p}) &=  \min \big\{L_{\bf r}(\mathbf{p} - {\bf r}, d_\mathbf{p}), 
    L_{\bf r} (\mathbf{p} -{\bf r}, d_\mathbf{p}-1) + P_1,\nonumber \\ 
    &L_{\bf r} (\mathbf{p} -{\bf r}, d_\mathbf{p}+1) + P_1, \min_{i} L_{\bf r}(\mathbf{p} - {\bf r}, i) +P_2 \big\}\nonumber \\
    & + C(\mathbf{p}, d_\mathbf{p}) - \min_k L_{\bf r} (\mathbf{p} - {\bf r}, k).
    \label{eq:sgm}
\end{align}
We obtain the optimal disparity of a pixel $d^{\ast}_\mathbf{p}$ by minimizing the aggregate cost along different paths as follows: 
\begin{align}
d^{\ast}_\mathbf{p} = \argmin_{d_\mathbf{p}} \sum_{\bf r} L_{\bf r}(\mathbf{p}, d_\mathbf{p}).
\label{eq:opt}
\end{align}
Finally, a parabola is fitted to the optimal disparities between the pixel and its two neighbors to obtain their subpixel disparities. 
\section{Methodology}
As shown in Fig.~\ref{fig:flow}, our approach consists of three main components: semidensification, SGM with DDC, and a stereo-LiDAR consistency check. 
The semidensification process generates a partially densified disparity map from the sparse disparity map $\bar{D}=\left\{\bar{d}_\mathbf{p} \in \mathbb{N} \cup \{\text{invalid}\} \mid \mathbf{p} \in \Omega \right\}$ using stereo images. 
We assume that $D$ and $\bar{D}$ are geometrically aligned and have the same pixel coordinates.
The DDC leverages a robust integration of stereo-LiDAR images to manage measurement noise and misprojection caused by occlusion and miscalibration. 
The stereo-LiDAR consistency check evaluates the consistency between the stereo images and LiDAR data. 

\subsection{Discrete Disparity-matching Cost}
\label{sec:method_cost}
We propose a disparity-matching cost that considers the sparse disparity map derived from LiDAR measurements into the proposed SGM framework. 
This matching cost applies penalties based on different scenarios and takes on discrete values, similar to the strategy used in SGM. 
LiDAR-SGM~\cite{forkel2023lidar} utilizes a quadratic cost; 
However, this approach tends to overpenalize when the disparity deviates significantly from prior disparity values and is less tolerant of misprojections in sparse disparity maps. 

We define DDC by the pepenalties $(0, Q_1, Q_2)$ for the following three cases: 
\begin{enumerate}
    \item $0$: no penalty if estimated disparity matches prior value; this preserves accurately measured data, 
    \item $Q_1$: small penalty when the estimated disparity slightly differs from the prior, thereby allowing for handling noise, 
    \item $Q_2$: fixed penalty for larger differences that accounts for misprojections and enables disparity estimation away from the prior, 
\end{enumerate}
under the condition that $Q_1 \leq Q_2$. 

The baseline stereo-matching cost is the Hamming distance of census-transformed images~\cite{spangenberg2013weighted}, $H\colon \Omega \times \mathbb{N} \to \mathbb{R}$. 
By combining the stereo-matching cost and DDC, we derive the joint-matching cost $\bar{C}$ as follows: 
\begin{align}
  \bar{C} (\mathbf{p}, d_\mathbf{p}) =& \left(1 - \alpha\right) H(\mathbf{p}, d_\mathbf{p}) 
  +  \alpha \bigg\{Q_1 T \left[ \left| d_\mathbf{p} - \bar{d}_\mathbf{p} \right| = 1 \right] \nonumber \\
  &+ Q_2 T \left[\left| d_\mathbf{p} - \bar{d}_\mathbf{p} \right| > 1\right]\bigg\},
  \label{eq:C}
\end{align}
where $\alpha$ is a parameter that balances the contributions of the stereo and disparity-matching costs. 
The cost $C$ in Eq.~\ref{eq:sgm} is replaced with $\bar{C}$ in Eq.~\ref{eq:C}, and we find the optimal disparity by solving Eq.~\ref{eq:opt}. 

Note that the sparse disparity map used for the DDC computation originates from either the semidensification step or directly from the LiDAR disparity data. 

\subsection{Semidensification}
\label{sec:method_semi}
Semidensification enhances the prior information extracted from the sparse disparity map to generate the semidense disparity map $\hat{D}=\left\{\hat{d}_\mathbf{p} \in \mathbb{N} \cup \{\text{invalid}\} \mid \mathbf{p} \in \Omega \right\}$. 
By increasing the density of the sparse disparity map, DDC gains more impact since DDC is only applied for pixels where the sparse disparity value exists. 
In addition, this process eliminates misprojections to ensure DDC robustness. 

The semidensification process fills the sparse disparity map at a pixel $\mathbf{p}$ with the semidense disparity $\hat{d}_\mathbf{p}$. 
This disparity, $\hat{d}_\mathbf{p}$, must minimize the cost of stereo matching within the disparities in $M_\mathbf{p}$, where $M_\mathbf{p}$ is defined as a window of size $(2r_s+1) \times (2r_s+1)$ centered at $\mathbf{p}$. 
In addition, the minimum matching cost must be less than the threshold value $T_s$. 
If no neighboring disparities meet these conditions, the sparse disparity value is used directly, and it remains invalid if no sparse data are available. 
Thus, the semidense disparity $\hat{d}_\mathbf{p}$ is calculated as follows: 
\begin{equation}
    \hat{d}_\mathbf{p} = \left\{ \begin{array}{ll} \argmin_{\bar{d}_\mathbf{q} | \mathbf{q} \in M_\mathbf{p}}{H\left(\mathbf{p}, \bar{d}_\mathbf{q}\right)} & {\rm if}\  \min{H\left(\mathbf{p}, \bar{d}_\mathbf{q}\right)} < T_s, 
    \\
    \bar{d}_\mathbf{p} & {\rm otherwise}. 
    \end{array}
    \right.
    \label{eq:semi}
\end{equation}
Note that this process may mitigate misprojections when the matching cost is high due to factors like occlusion by replacing the disparity value with a neighboring value that achieves a low matching cost. 
The semidense disparity $\hat{d}_\mathbf{p}$ is used as an alternative to the original sparse disparity $\bar{d}_\mathbf{p}$ when the matching cost is calculated (Eq.~\ref{eq:C}). 

We refrain from applying spatial smoothing during the semidensification process because the addition of a smoothness term would lead to high computational costs. 
In addition, preserving the details of small or thin objects is more effective at this stage. 
Because spatial smoothing is applied later in the SGM process, the semidensification step focuses exclusively on the matching cost.

\subsection{Stereo-LiDAR consistency check}
\label{sec:method_consistency}
The consistency check filters out disparity values that are not consistent across multiple views. 
In SGM~\cite{hirschmuller2007stereo}, the consistency check uses two camera views. 
However, stereo-LiDAR fusion involves three views, two from the cameras and one from the LiDAR; thus, it is more efficient to incorporate the LiDAR view into the consistency check process than to rely solely on stereo camera views. 

Assume that we estimate the disparity $d_\mathbf{p}^{\ast}$ at a pixel $\mathbf{p}$ of the base image and the disparity of the matching pixel in the other image $d_{\mathbf{q}_m}^{\prime}$. 
The pixel $\mathbf{q}_m$ is obtained by traversing the epipolar line on the matching image: $\mathbf{q}_m = e_{\rm bm}(\mathbf{p}, d^{\ast}_\mathbf{p})$. 
If the corresponding disparities differ significantly, the disparity is set to invalid~\cite{hirschmuller2007stereo}. 
The proposed method includes a consistency check between the base camera and LiDAR. 
We consider that the disparity is consistent if the estimated disparity $d^{\ast}_\mathbf{p}$ matches at least one of the disparities in its neighboring pixels in the sparse disparity map $\bar{d}_{\mathbf{q}}\ |\ \mathbf{q}\in K_\mathbf{p}$. 
Here, $K_\mathbf{p}$ is defined as a window of size $(2r_c + 1) \times (2r_c + 1)$ centered at $\mathbf{p}$. 
$d^{\ast}_\mathbf{p}$ and $\bar{d}_\mathbf{q}$ are considered to be matched if $\left|d^{\ast}_\mathbf{p} - \bar{d}_\mathbf{q}\right| \leq T_c$, where $T_c$ is a given threshold value. 
Finally, the integrated three-view consistency check can be expressed as follows: 
\begin{equation}
    d^{c \ast}_\mathbf{p} = \left\{ \begin{array}{ll}d^{\ast}_{\mathbf{p}} & \begin{array}{l}
    {\rm if} \left|d^{\ast}_\mathbf{p} - d^{\prime}_{\mathbf{q}_m} \right|\leq 1, \mathbf{q}_m = e_{\rm bm}(\mathbf{p}, d^{\ast}_\mathbf{p}), \\[2pt]
    {\rm or\ if} \min_{\mathbf{q} \in K_\mathbf{p}} {\left|d^{\ast}_\mathbf{p} - \bar{d}_\mathbf{q}\right|} \leq T_c. \\[2pt]
    \end{array}\\
    {\rm invalid} & \begin{array}{l} {\rm otherwise}. \\[2pt]\end{array}
    \end{array}\right.
    \label{eq:consistency}
\end{equation}
Note that we did not use the semidense disparity map during the consistency check because the propagated disparities may not satisfy the geometric relations between the sensors. 
The final output of our method is a consistent dense disparity map $D^{c\ast}=\left\{d^{c\ast}_\mathbf{p}  \mid \mathbf{p} \in \Omega\right\}$ derived by Eq.~\ref{eq:consistency}. 

\begin{table}[t]
    \centering
    \caption{Variations of our method in the evaluations}
    \begin{tabular}{c|c| c| c}
    \hline
        \multirow{2}{*}{Name} & Semi- & SGM with & Stereo-LiDAR  \\
        & densification & DDC & consistency check \\
        \hline
         DSGM & &$\surd$&$\surd$\\
         SDSGM &$\surd$&$\surd$&$\surd$\\
         \hline
    \end{tabular}
    \label{tab:ours}
\end{table}

\begin{table}[t]
    \centering
    \caption{Our parameters in the evaluations}
    \begin{tabular}{c|c|c|c}
    \hline
         Stage & Name &Value & Meaning\\
         \hline
         Semi-&$T_s$ & 2 & Threshold for semidensification\\
         densification&$r_s$& 6 & Window size for semidensification\\
         \hline
         &$P_1$ & 10 & Small SGM smoothness cost\\
         SGM&$P_2$ & 120 & Large SGM smoothness cost\\
         with&$Q_1$ & 5 & Small disparity-matching cost\\
         DDC&$Q_2$ & 160 & Large disparity-matching cost\\
         &$\alpha$ & 0.7 & Blending ratio of costs\\
         \hline
         Consistency-&$T_c$ & 2 & Threshold for consistency check\\
         check& $r_c$&20&Window size for consistency check\\
    \hline
    \end{tabular}
    \label{tab:param}
\end{table}

\begin{table*}[t]
\centering
\caption{Disparity estimation results on KITTI 141}
\label{tab:eval_kitti}
\begin{tabular}{c|c|c|ccc|cc|c}
\hline
\multirow{2}{*}{Method}
& \multirow{2}{*}{Input} & \multirow{2}{*}{Non-learning} & \multirow{2}{*}{Realtime} & Time& GPU & Coverage& Covered& Total\\
& & & & [ms] & platform & [\%] & error [\%] & error [\%]\\

\hline
    
SGM~\cite{hirschmuller2007stereo} & Stereo & $\surd$ & $\surd$ & 37 &Jetson Orin NX& 93.0 & 3.73 & 6.00  \\

JointEst.~\cite{yamaguchi2014efficient} & Stereo & $\surd$ &  & 1947 & - & 99.9 & 4.32 & 4.33 \\
     
SSM-TGV~\cite{yao2021non} & Stereo + LiDAR & $\surd$ &  & 250 &Jetson Orin NX&  100.0 & 3.32 & 3.32  \\   

Probabilistic~\cite{maddern2016real} & Stereo + LiDAR & $\surd$ & $\surd$ & (24) &AMD R9 295x2&  (99.6) & (5.91) & - \\

LiDAR-SGM~\cite{forkel2023lidar} & Stereo + LiDAR & $\surd$ & $\surd$ & (24) &GTX 1050 Ti& (97.5) & (3.87) &  - \\

LiDAR-SGM~(Huber) & Stereo + LiDAR & $\surd$ & $\surd$ & 39 &Jetson Orin NX& 93.7 & 3.50 &  4.93 \\

DSGM~(Ours) & Stereo + LiDAR & $\surd$ & $\surd$ & 40 &Jetson Orin NX& 99.0 & 2.81 & 3.15\\

SDSGM~(Ours) & Stereo + LiDAR & $\surd$ & $\surd$ & 50 &Jetson Orin NX& 99.6 & {\color{red}\textbf{2.61}} & {\color{red}\textbf{2.79}}\\

\hline
CNN~\cite{park2018high} & Stereo + LiDAR &  & $\surd$ & (45) &Titan X& (99.8) & (4.84) & - \\

Fastfusion~\cite{meng2023fastfusion} & Stereo + LiDAR&  &  $\surd$ & (49) & Titan Xp& 100.0 & 3.05 & 3.05 \\

CCVNorm~\cite{wang20193d} & Stereo + LiDAR && & (1011) &GTX 1080 Ti& (100.0) & (3.35) & (3.35) \\

LSNet~\cite{cheng2019noise} & Stereo + LiDAR& &  & 3284&Jetson Orin NX&100.0 & {\color{blue} \textbf{2.17}} & {\color{blue} \textbf{2.17}}\\
     \hline
     \multicolumn{9}{c}{The {\color{red}\textbf{bests among non-learning or realtime methods}} are indicated in red. The {\color{blue}\textbf{bests among all}} are indicated in blue.}\\
     \multicolumn{9}{c}{Values in brackets were obtained from the cited papers.}\\
\end{tabular}
\end{table*}

\begin{figure*}[t]
\centering
\includegraphics[width=1.0\textwidth]{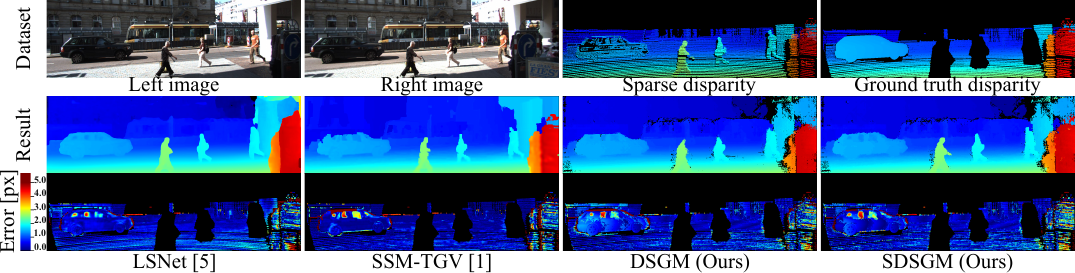}
\caption{Dataset and results of KITTI 141 evaluation. 
Overall, offline learning-based LSNet~\cite{cheng2019noise} showed the least error, as seen on the car in the error maps. 
SSM-TGV~\cite{yao2021non} showed a more significant error than our methods, as seen on the car roof in the error map.}
\label{fig:result_kitti}
\end{figure*}

\section{Experiments}
\label{sec:eval}
We compared the proposed method to SOTA stereo-LiDAR fusion and non-learning stereo approaches by analyzing the contribution of each component and its robustness across different scenarios. 
Our evaluation covers the accuracy of and processing time required by the proposed method (Sec.~\ref{sec:eval_bench}), the impact of semidensification, and the role of the consistency check (Sec.~\ref{sec:eval_ablation}). 
In addition, we tested the proposed method's robustness under various conditions, including different input densities (Sec.~\ref{sec:eval_density}), other weather conditions, and indoor scenes (Sec.~\ref{sec:eval_domain}). 
We also evaluated the effect of varying parameters (Sec.~\ref{sec:param}). 
Ablation studies highlight the differences in performance between the DSGM and SDSGM variations, as detailed in Table~\ref{tab:ours}, where the distinction between the variations is the application of semidensification. 
The parameters used in our evaluations other than Sec.~\ref{sec:param} are described in Table~\ref{tab:param}. 

\subsection{Implementation and dataset}
\label{sec:dataset}
The proposed method was integrated with an open-source SGM implementation in CUDA\footnote{https://github.com/fixstars/libSGM}. 
The comparative methods were run on our platform when the implementation was available. 
Otherwise, we referenced the results reported in the original studies. 
The platform used in these experiments was an NVIDIA Jetson Orin NX with 16GB of memory. 
The source code of the proposed method is available at \href{https://github.com/yshry/libSGM_lidar}{https://github.com/yshry/libSGM\_lidar}.

We utilized the KITTI 141 dataset, which is a subset of the KITTI stereo dataset~\cite{geiger2012we}. 
The KITTI 141 was extracted by~\cite{maddern2016real} and is one of the datasets most commonly used to benchmark stereo-LiDAR fusion methods. 
The dataset contains 141 sets of rectified stereo images, LiDAR point clouds captured by Velodyne HDL 64E, and corresponding ground truth dense disparity images (Sec.~\ref{sec:eval_bench} and \ref{sec:eval_ablation}). 
We also used 32- and 16-scan-line disparity maps created by vertically sampling the 64-scan-line original map to half and quarter densities based on scan angle (Sec.~\ref{sec:eval_density}). 
For the evaluation, we used the code provided with the KITTI benchmark\footnote{https://www.cvlibs.net/datasets/kitti/eval\_scene\_flow.php}. 

To evaluate the adaptability to various scenes (Sec.~\ref{sec:eval_domain}), we applied the method to the CARLA\footnote{https://www.mucar3.de/icra2023-lidar-sgm} and Middlebury\footnote{https://vision.middlebury.edu/stereo/data/scenes2021/} dataset. 
CARLA is a dataset of simulated outdoor scenes proposed in~\cite{forkel2023lidar}, including 500 sets of rectified stereo images and 64-scan-lines LiDAR data under two coupled weather and time conditions, named "ClearSunset" and "HardRainNoon". 
Middlebury is a dataset of indoor scenes by~\cite{scharstein2014high}. 
We utilized the 2021 mobile dataset, including 24 sets of rectified stereo images and a ground truth disparity map. 
We randomly sampled 1\% of each ground truth map to obtain the corresponding sparse disparity map. 

\subsection{Overall performance}
\label{sec:eval_bench}
First, we compare the overall performance of the proposed method with the performances of existing approaches. 
In addition, to evaluate DDC compared with an outlier-robust cost, we implemented a variant of LiDAR-SGM~\cite{forkel2023lidar} using Huber cost. 
The KITTI evaluation code provides two error rates: the covered error and the total error. 
The covered error measures the error only in regions where valid estimations exist, excluding invalid areas. 
In contrast, the total error calculates the error rate by filling invalid pixels through the background interpolation~\cite{hirschmuller2007stereo}. 
We used the error rate to represent the percentage of cases where the estimated value differed from the ground truth by three pixels or more. 
Because most LiDARs work at 10--20 Hz, we consider a method to be real-time if its processing time per frame is less than 100 ms. 

Table~\ref{tab:eval_kitti} shows the quantitative evaluation results. 
Among the non-learning methods, the proposed method achieves sufficient coverage and the lowest error rate for both the covered and total errors. 
In addition, the proposed method outperforms methods that are not real-time. 
By comparing the results of DSGM with LiDAR-SGM~\cite{forkel2023lidar} and its variant, we found DDC was more effective than the quadratic and Huber costs. 
We consider this because DDC's discreteness managed the misprojection and noise in the sparse disparity map more efficiently than other costs. 
Due to the high coverage rate of learning-based methods, directly comparing the covered error rates does not constitute a fair comparison. 
However, the total error rate of the proposed method is lower than that of FastFusion~\cite{meng2023fastfusion}. 
Although LSNet~\cite{cheng2019noise} achieves the lowest overall error rate, it requires approximately 60 times more computational time than does the proposed approach. 

\begin{figure}[t]
\centering
    \begingroup
    \captionsetup{justification=centering}
    \begin{minipage}[t]{0.18\hsize}
    \includegraphics[width=1.0\textwidth]{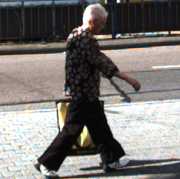}
    
    \vspace{2mm}
    \includegraphics[width=1.0\textwidth]{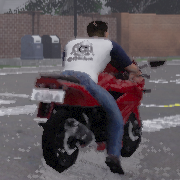}
    \subcaption{Images}
    \end{minipage}
    \begin{minipage}[t]{0.18\hsize}
    \includegraphics[width=1.0\textwidth]{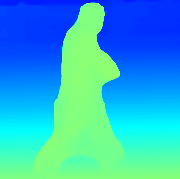}
    
    \vspace{2mm}
    \includegraphics[width=1.0\textwidth]{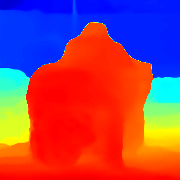}
    \subcaption{LSNet\cite{cheng2019noise}}
    \end{minipage}
    \begin{minipage}[t]{0.18\hsize}
    \includegraphics[width=1.0\textwidth]{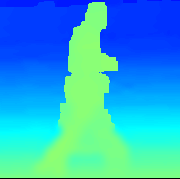}
    
    \vspace{2mm}
    \includegraphics[width=1.0\textwidth]{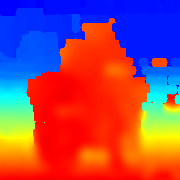}
    \subcaption{SSM-\\TGV\cite{yao2021non}}
    \end{minipage}
    \begin{minipage}[t]{0.18\hsize}
    \includegraphics[width=1.0\textwidth]{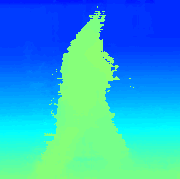}
    
    \vspace{2mm}
    \includegraphics[width=1.0\textwidth]{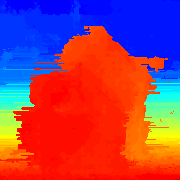}
    \subcaption{DSGM\\(Ours)}
    \end{minipage}
    \begin{minipage}[t]{0.18\hsize}
    \includegraphics[width=1.0\textwidth]{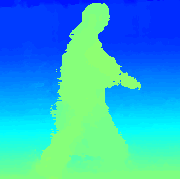}
    
    \vspace{2mm}
    \includegraphics[width=1.0\textwidth]{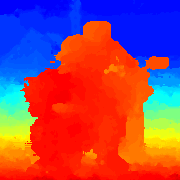}
    \subcaption{SDSGM\\(Ours)}
    \end{minipage}
    
    \endgroup
    \caption{Resulting SDSGM silhouettes are finer than those of the other methods (Upper: KITTI and Lower: CARLA HardRainNoon). 
    A background interpolation~\cite{hirschmuller2007stereo} was used to carry out the visual comparison. 
    The effect is visible at (Upper) the arm and (Lower) the side mirror areas.}
    \label{fig:silhouette}
\end{figure}

Figure~\ref{fig:result_kitti} shows the example dataset and visual results, highlighting the qualitative evaluations. 
The figure shows the best methods of non-learning and learning for comparison. 
As in Fig.~\ref{fig:silhouette}, SDSGM successfully recovered the detailed silhouette of the pedestrian. 
Conventional methods often oversmooth thin or small objects due to their strong constraints on smoothness. 
In contrast, the semidensification effectively generates the proper prior information for these objects because it does not enforce smoothness terms. 

\begin{table}[t]
    \centering
    \caption{Semidensification effects on KITTI 141}
    \begin{tabular}{c|cc}
    \hline
    &Coverage [\%] & Covered error [\%] \\
    \hline
         Input sparse disparity map &  7.3 & 4.66\\
         Semidense disparity map& 28.6 & \textbf{4.15} \\
    \hline
    \end{tabular}
    \label{tab:semidensification}
\end{table}

\begin{figure}[t]
\centering
    \centering
    \begin{minipage}[t]{0.30\hsize}
    \includegraphics[width=1.0\textwidth]{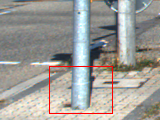}
    
    \vspace{2mm}
    \includegraphics[width=1.0\textwidth]{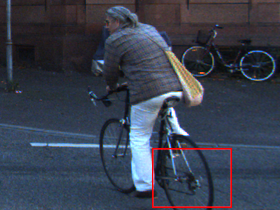}
    \subcaption{Image}
    \end{minipage}
    \begin{minipage}[t]{0.30\hsize}
    \includegraphics[width=1.0\textwidth]{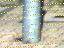}

    \vspace{2mm}
    \includegraphics[width=1.0\textwidth]{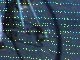}
    
    \subcaption{Sparse map}
    \end{minipage}
    \begin{minipage}[t]{0.30\hsize}
    \includegraphics[width=1.0\textwidth]{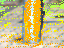}
    
    \vspace{2mm}
    \includegraphics[width=1.0\textwidth]{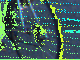}
    \subcaption{Semidense map}
    \end{minipage}

    \caption{Top-row images: Misprojections in (b) the sparse map appear as disparities of the road (yellow dots) on the pole.
    Misprojections decreased in (c) the semidense disparity.
    Bottom-row images: (b) A few LiDAR disparities are projected on thin objects, such as the bicycle wheels.
    (c) The semidense map contains disparity values for such objects.}
    \label{fig:semidensification}
\end{figure}

\subsection{Ablation studies}
\label{sec:eval_ablation}
\subsubsection{Semidensification}
\label{sec:eval_semi}
We verified the impact of semidensification in enhancing sparse input disparities in SGM with DDC. 
Table~\ref{tab:semidensification} compares the original sparse disparity maps with our semidense disparity maps on the KITTI 141 dataset in terms of accuracy and density. 
The results indicate that semidensification improves both the coverage and error rates. 
These improvements are visually demonstrated in Fig.~\ref{fig:semidensification}, where semidensification simultaneously removes misprojection and densifies the disparity map. 

\begin{table}[t]
    \centering
    \caption{Consistency check effects on KITTI 141}
    \begin{tabular}{c|cc|c}
    \hline
    Input &Coverage [\%] & Covered error [\%] & Total error [\%]\\
    \hline
         Stereo &97.1 & \textbf{1.95}&3.12\\
         Camera-LiDAR & 99.4& 2.57&2.81 \\
         Stereo-LiDAR &99.6&2.61&\textbf{2.79} \\
    \hline
    \end{tabular}
    \label{tab:consistency}
\end{table}

\begin{figure}[t]
\centering
    \centering
    \begin{minipage}[t]{0.3\hsize}
    \includegraphics[width=1.0\textwidth]{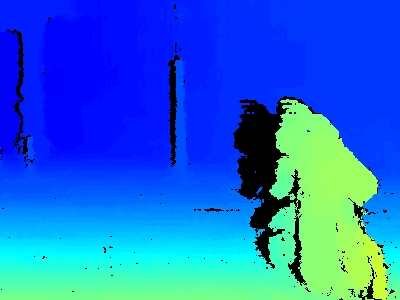}
    \subcaption{Stereo}
    \end{minipage}
    \begin{minipage}[t]{0.3\hsize}
    \includegraphics[width=1.0\textwidth]{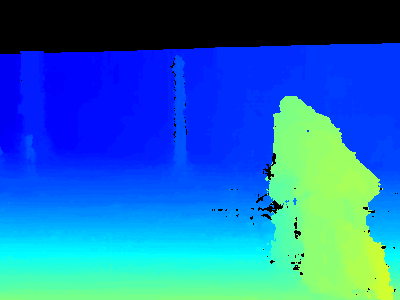}
    \subcaption{Camera-LiDAR}
    \end{minipage}
    \begin{minipage}[t]{0.3\hsize}
    \includegraphics[width=1.0\textwidth]{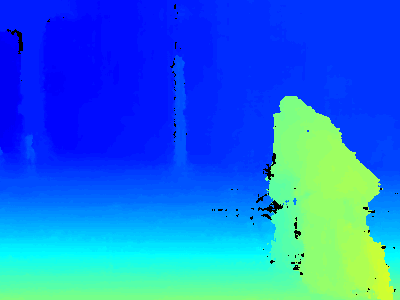}
    \subcaption{Stereo-LiDAR}
    \end{minipage}
    \caption{Qualitative evaluations of the consistency checks.
    (a) The stereo check labels the area blocked in the second camera as invalid.
    (b) Camera-LiDAR check labels the area outside the LiDAR field of view as invalid.
    (c) Stereo-LiDAR check has the most valid pixels.}
    \label{fig:consistency}
\end{figure}

\subsubsection{Stereo-LiDAR consistency check}
\label{sec:eval_consistency}
We evaluated the proposed stereo-LiDAR consistency check and compared it with the conventional stereo consistency check~\cite{hirschmuller2007stereo} and the camera-LiDAR consistency check. 
Here, the camera-LiDAR check only checks the consistency between the base camera and LiDAR. 
The quantitative and qualitative results are shown in Table~\ref{tab:consistency} and Fig.~\ref{fig:consistency}, in which all disparity maps have been generated by SDSGM prior to the consistency checks. 
Since the stereo and camera-LiDAR consistency checks label more invalid pixels than the stereo-LiDAR check, they reduce coverage and improve the covered error. 
Meanwhile, the stereo-LiDAR check achieved the most coverage and the least total error. 

\subsection{Robustness to LiDAR density}
\label{sec:eval_density}
We applied the proposed and comparison methods to KITTI 141 with 32 and 16 LiDAR scan lines and obtained the results in Table~\ref{tab:input_variation}. 
The proposed method outperformed a previous non-learning offline SOTA approach~\cite{yao2021non} for maps having both 32 and 16 scan lines. 
In addition, with maps having 32 scan lines, the proposed method achieved a total error of 3.27\% and outperformed the result of~\cite{yao2021non} with maps having 64 scan lines, which was 3.32\% (refer to Table~\ref{tab:eval_kitti}). 

\begin{table}[t]
    \centering
    \caption{Results on different LiDAR densities}
    \begin{tabular}{c|c|c|cc|c}
        \hline
        Scan& \multirow{2}{*}{Method} &Time & Cove-& Covered& Total\\
        lines& &[ms] & rage[\%] &error[\%] &error[\%]\\
        \hline
         & SSM-TGV~\cite{yao2021non} & 241 &100.00& 4.00 & 4.00 \\
         32& DSGM~(Ours) & 39 &96.82 & 3.04 &  3.77\\
         & SDSGM~(Ours) & 51 &98.99 & \textbf{2.92} & \textbf{3.27} \\
        \hline
         & SSM-TGV~\cite{yao2021non} & 239 & 100.00 & 5.14 & 5.14 \\
         16& DSGM~(Ours)& 37 & 94.79 & 3.32 & 4.43 \\ 
         & SDSGM~(Ours)& 51 & 98.11 & \textbf{3.34} & \textbf{3.91} \\
        \hline
    \end{tabular}
    \label{tab:input_variation}
\end{table}

\begin{table}[t]
    \centering
    \caption{Results on various datasets}
    \begin{tabular}{c|c|c|cc|c}
        \hline
        \multirow{3}{*}{Dataset} &\multirow{3}{*}{Method} & Time&Cove-& Cover-& Total\\
        & && rage &ed err-& error\\
        & &[ms]&[\%] &or[\%]&[\%]\\
        \hline
         &SGM~\cite{hirschmuller2007stereo} &33& 69.5 & 11.20 & 21.58 \\
         CARLA&JointEst.~\cite{yamaguchi2014efficient} &1824& 99.9 & 10.21 & 10.26  \\
         Clear&SSM-TGV~\cite{yao2021non} &239& 99.4 & 12.97 & 13.10 \\
         Sunset& LSGM~\cite{forkel2023lidar}&-& - & - & (12.95) \\
         (Outdoor)&DSGM~(Ours)  &35& 80.9&\textbf{7.34} &10.29\\
         &SDSGM~(Ours)  &49& 87.5 & 7.40 & \textbf{10.05}\\
         \cline{2-6}
         &LSNet~\cite{cheng2019noise}  &3123&100.0 & 17.26 & 17.26\\
         \hline
         &SGM~\cite{hirschmuller2007stereo} &34& 63.6 & 18.77 & 34.45\\
         CARLA&JointEst.~\cite{yamaguchi2014efficient} &1834& 99.6 & 23.39 & 23.58\\
         Hard&SSM-TGV~\cite{yao2021non}  &239& 99.4 &13.35 & 13.49 \\
         Rain& LSGM~\cite{forkel2023lidar} &-& - & - & (14.03) \\
         Noon&DSGM~(Ours) &35& 74.0 & \textbf{8.94}  & \textbf{12.16} \\
         (Outdoor)&SDSGM~(Ours)  &52& 77.0 &9.30 & 12.34 \\
         &$\dagger$\textit{SDSGM~(Ours)} & \textit{54} & \textit{92.6} & \textit{8.23} & \textit{10.94} \\
         \cline{2-6}
         &LSNet~\cite{cheng2019noise} &3124& 100.0 & 28.69 & 28.69 \\
         \cline{2-6}
         \hline
         &SGM~\cite{hirschmuller2007stereo}&38&81.3&10.00&16.78\\
         Middle-&JointEst.\cite{yamaguchi2014efficient}&1423&99.9&13.50&13.45 \\
         burry&SSM-TGV~\cite{yao2021non} &217&99.7&15.80&15.80 \\
         (Indoor)&DSGM~(Ours) &43&92.8&7.55&9.43\\
         &SDSGM~(Ours) &59&95.8&\textbf{6.17}&\textbf{7.30}\\
         \cline{2-6}
         &LSNet~\cite{cheng2019noise}&4621&100.0&10.46&10.46\\
         \hline
         \multicolumn{6}{c}{$\dagger$ A reference using proper semidensification parameter ($T_s=9$).}\\
         \multicolumn{6}{c}{Values in brackets were obtained from the cited papers.}\\
    \end{tabular}
    \label{tab:carla}
\end{table}

\begin{figure*}[t]
\centering
\includegraphics[width=1.0\hsize]{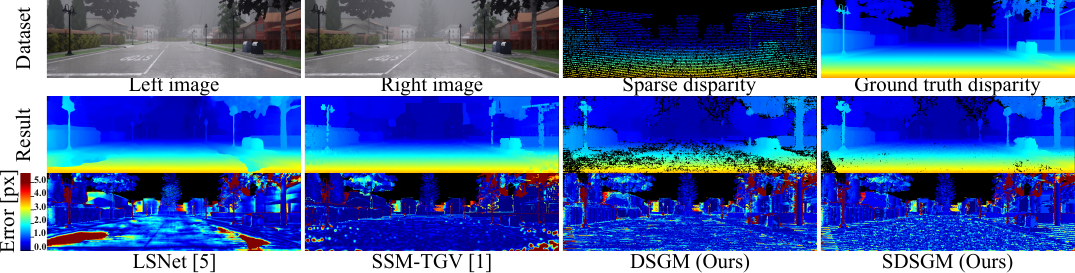}
\caption{CARLA HardRainNoon dataset and results. 
The displayed SDSGM result is obtained using a proper semidensification threshold ($\dagger$ in Table~\ref{tab:carla}). 
LSNet~\cite{cheng2019noise} showed a significantly larger error than non-learning methods. 
SSM-TGV\cite{yao2021non} showed the significant error at the right lowest corner of the image. 
Overall, our methods showed more minor errors than the compared methods.}
\label{fig:results_carla}
\end{figure*}

\begin{figure*}
\centering
\includegraphics[width=1.0\hsize]{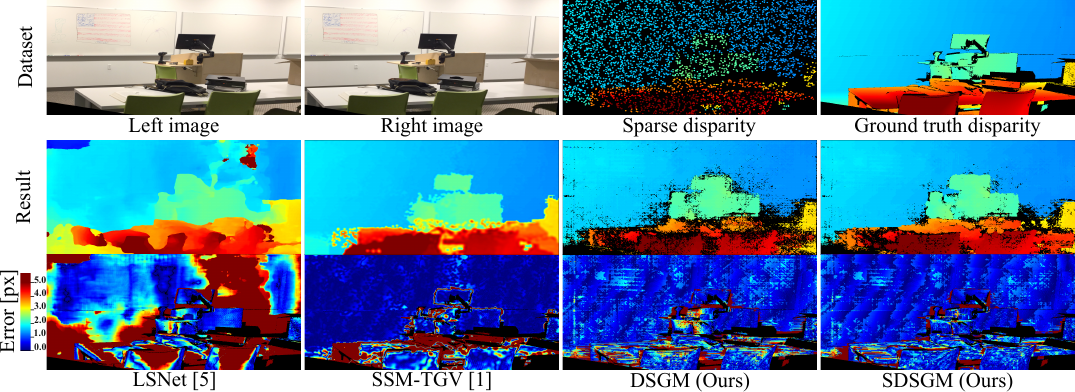}
\caption{Middlebury dataset and results. 
LSNet~\cite{cheng2019noise} showed a significantly larger error than non-learning methods. 
SSM-TGV~\cite{yao2021non} has a significant error on the desk board. 
Overall, our methods showed more minor errors than the compared methods.}
\label{fig:results_middlebury}
\end{figure*}

\subsection{Adaptability to various datasets}
\label{sec:eval_domain}
To assess the performance of the proposed method across different scenarios, we evaluated the method by CARLA ClearSunset, CARLA HardRainNoon, and Middlebury datasets. 
For adaptability evaluation purposes, the parameters used in the compared methods and ours were the same as those used in the KITTI experiments. 
Figure~\ref{fig:results_carla} and \ref{fig:results_middlebury} presents the visual results, and Table~\ref{tab:carla} provides the quantitative results. 
The proposed SDSGM achieved the best performance on the ClearSunset and Middlebury datasets. 
LSNet, which utilized the same parameters and NN weights as in Sec.~\ref{sec:eval_bench}, showed a significantly higher error rate than did the non-learning methods. 
This demonstrates the advantage of non-learning methods over learning-based approaches in terms of adaptability without fine-tuning. 

For the HardRainNoon dataset, DSGM outperformed SDSGM, suggesting that semidensification does not improve the accuracy in this scenario. 
In contrast, when using a proper parameter ($T_s=9$), semidensification enhances accuracy, as in the row indicated with $\dagger$ in Table~\ref{tab:carla}. 
Hard rain scenes present a challenge to stereo vision because rain introduces noise into the image, which increases the stereo-matching cost, even for correct disparity estimations. 
As a result, $T_s$ should be higher than in other scenarios to accommodate the larger stereo-matching cost.

\subsection{Parameter Study}
\label{sec:param}
\begin{figure*}[t]
\centering
\includegraphics[width=1.0\linewidth]{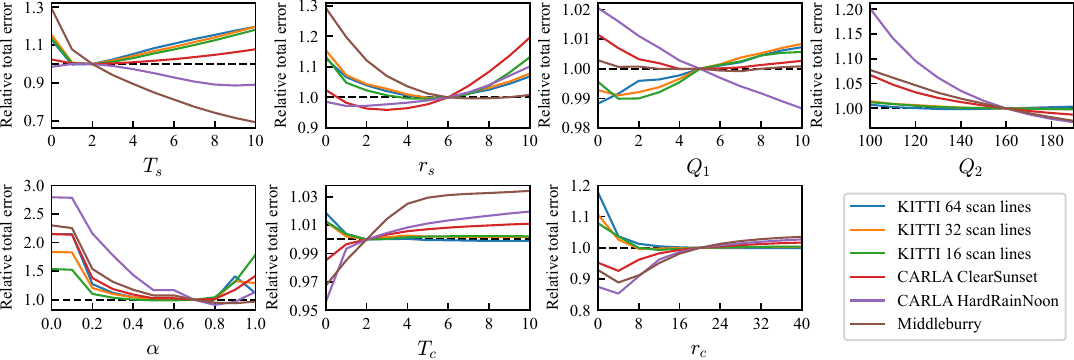}
\caption{Parameter study results. 
In a graph, the X-axis is the varied parameter, and the Y-axis is the total error relative to the total error using the parameters in Table~\ref{tab:param}. 
The parameter-wise observations follow. 
Large $T_s$ and $r_s$, which densify the semidense disparity maps, improved the results for Middlebury. 
We consider this due to the less coverage of Middlebury sparse disparity maps (1 \%) than other datasets. 
Large $Q_1$ and $Q_2$, which strongly impose the sparse disparity matching constraints, performed well for CARLA and Middlebury. 
We consider this to be the case because the sparse data in these artificial datasets (created through computer graphics or sampling) is more reliable than in real datasets. 
Similarly, small $T_c$ and $r_c$, which lead to the consistency check to the stereo-only like, improved results for CARLA and Middlebury. 
The reason for this is considered as follows. 
Since the LiDAR is located at the same position as the left camera in these artificial datasets, the LiDAR consistency check is biased to keep more foreground disparities and less background disparities. 
In contrast, for the real dataset of KITTI, we see the stereo-LiDAR consistency check improved the results.}
\label{fig:param}
\end{figure*}

We evaluated the effect of parameters across all datasets used above. 
We used the same values of $P_1$ and $P_2$ as the original stereo SGM implementation. 
For other parameters we introduced, we evaluated the effect on the total error by varying them. 
The graphs and discussions in Fig.~\ref{fig:param} highlight the evaluation. 
The effects of each parameter were relatively independent, so each graph in Fig.~\ref{fig:param} focuses on one parameter for clarity of illustration. 
Although parameter tuning is possible for each dataset, we found that the parameters in Table~\ref{tab:param} are well-balanced for different scenarios.

\section{Conclusion}
\label{sec:conclusion}
Stereo-LiDAR fusion is a technology that enhances depth estimation by combining stereo matching with LiDAR data. 
We focus on real-time, non-learning stereo-LiDAR fusion because it can be applied across various domains without the need for additional network training. 
The proposed method integrates SGM with DDC, semidensification, and a stereo-LiDAR consistency check. 
We demonstrate that the proposed method achieves improved performance over previous real-time stereo-LiDAR fusion methods in terms of error rate and demonstrates strong adaptability across different domains.

\bibliography{bib}
\bibliographystyle{IEEEtran}

\end{document}